\pdfoutput=1

\documentclass[11pt]{article}

\usepackage{acl}

\usepackage{times}
\usepackage{latexsym}

\usepackage[T1]{fontenc}

\usepackage[utf8]{inputenc}

\usepackage{microtype}

\usepackage{inconsolata}
\usepackage{subfiles}

\usepackage{graphicx}
\usepackage{xcolor}
\usepackage{multirow}
\usepackage{arydshln}
\usepackage{booktabs}
\usepackage{nicematrix}
\usepackage{makecell}

\usepackage{tabularx}
\usepackage{array}
\usepackage{xcolor,soul}
\usepackage{duckuments}
\usepackage{graphicx}
\usepackage{pifont}
\usepackage{arydshln}
\usepackage{amssymb}
\usepackage{amsmath}
\usepackage{cleveref}
\usepackage{enumitem}
\usepackage{hyperref}
\usepackage[normalem]{ulem}
\title{Improving Event Definition Following For Zero-Shot Event Detection}

\author{
Zefan Cai$^{*\dagger}$, 
Po-Nien Kung$^{*\ddagger}$, 
Ashima Suvarna$^\ddagger$, 
Mingyu Derek Ma$^\ddagger$, 
\textbf{Hritik Bansal}$^\ddagger$, \\
\textbf{Baobao Chang}$^\dagger$, 
\textbf{P. Jeffrey Brantingham}$^\ddagger$, 
\textbf{Wei Wang}$^\ddagger$, 
\textbf{Nanyun Peng}$^\ddagger$
\\
$^\dagger$Peking University,  
$^\ddagger$University of California, Los Angeles \\
\texttt{Zefncai@gmail.com, ponienkung@cs.ucla.edu}
}

\begin{document}
\maketitle
\renewcommand{\thefootnote}{\fnsymbol{footnote}}
\footnotetext[1]{Equal contribution.}

\begin{abstract}
Existing approaches on zero-shot event detection usually train models on datasets annotated with known event types, and prompt them with unseen event definitions. These approaches yield sporadic successes, yet generally fall short of expectations.
In this work, we aim to improve zero-shot event detection by training models to better follow event definitions. We hypothesize that a diverse set of event types and definitions are the key for models to learn to follow event definitions while existing event extraction datasets focus on annotating many high-quality examples for a few event types. 
To verify our hypothesis, we construct an automatically generated Diverse Event Definition (DivED) dataset and conduct comparative studies. Our experiments reveal that a large number of event types (200) and diverse event definitions can significantly boost event extraction performance; on the other hand, the performance does not scale with over ten examples per event type.
Beyond scaling, we incorporate event ontology information and hard-negative samples during training, further boosting the performance.
Based on these findings, we fine-tuned a LLaMA-2-7B model on our DivED dataset, yielding performance that surpasses SOTA large language models like GPT-3.5 across three open benchmarks on zero-shot event detection.


\looseness=-1

\end{abstract}

\section{Introduction}

\begin{figure}
    \centering
    \includegraphics[width=\columnwidth]{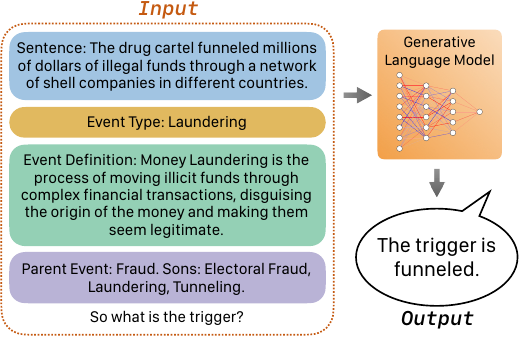}
    \caption{
    Zero-shot generative event detection formulation. We demonstrate a generated event type and sample from our DivED dataset. The input prompt includes information about \textit{Event Type}, \textit{Event Definition}, \textit{Event Ontology} and the query passage, and the expected output is a verbalized extracted result. 
    }
    \vspace{-1.5em}
    \label{fig:teaser}
\end{figure}

Event detection (ED) focuses on identifying event triggers of specific event types in a given text with predefined event ontology. 
Prior work has studied event detection largely in a fully-supervised fashion~\cite{wadden2019entity,lin2020joint,nguyen2015event,nguyen2016joint,han2019deep,du2020event,huang2020biomedical,huang2020document,paolini2021structured,ma-etal-2023-dice}. While these work show promising performance on seen events, it cannot generalize well to long-tailed and unseen events ~\cite{ma2023star,zhang2022efficient}.
To further enable generalization to low-resource events, prior work proposed to tackle few-shot event detection by training model on generated pseudo data~\cite{ma2023star, Kumar2020DataAugmentationUsing, Schick2021GeneratingDatasetsPretrained}. 
Despite the success in data-efficient event detection, they cannot zero-shot extract unseen events in real-time due to the need for prior training, limiting their applicability to a wider range of scenarios.\looseness=-1

The success of task generalization of LLMs enabled by instruction tuning further advances zero-shot event detection.
Recent work started to extract events of novel type by providing LLMs with the event definition of unseen events during inference, as demonstrated in \autoref{fig:teaser}. They either prompt closed-source LLMs, such as GPT-3.5
~\cite{wang2022code4struct,gao2023exploring,wei2023zero}, or apply transfer learning on open-sourced LMs with EE training data of seen event types~\cite{huang-etal-2018-zero,lyu2021zero}. While the former methods achieve acceptable performance, they are not flexible and reproducible due to their closed nature, leading to the difficulty in further improving the models' performance. In contrast, the latter methods while reproducible and flexible, suffer from low performance.\looseness=-1

In this work, we aim to enhance zero-shot event detection (ED) by training a model with improved generalization to unseen event types. During inference, the model, prompted with definitions of previously unseen events, relies on its instruction-following ability to understand event definitions and identify correct triggers. Despite recent impressive results of instruction following by Large Language Models (LLMs), there is room for improvement~\cite{kung-peng-2023-models, yin2023did}, and we focus on enhancing it via instruction fine-tuning with strategically generated data. 
Specifically, we hypothesize that transfer learning from conventional EE datasets might not be ideal. 
Though a large amount of high-quality training samples for only a few event types equips the model to perform EE on homogeneous data, it is not sufficient for the model to develop generalizability towards unseen situations.
Different from existing works that aim to improve ED model by generating more homogeneous EE data~\cite{ma2023star}, we posit that a large number of event types and a diverse set of event definitions are the keys to improving the event definition following capabilities.\looseness=-1

To verify our assumption, 
we develop \textbf{Diverse Event Definition (DivED)} Dataset, which is generated from LLMs with diverse event definitions and samples. DivED includes 3000+ event types, each with 10 event definitions and 10 samples. Since event types can be organized into tree-structure ontology, we further inject each event's event type dependencies information into its event definition, including the name of its parent and sibling events. \looseness=-1


Our study on the DivED dataset supports our hypothesis. The results indicate that a sufficient amount of event types (200) and diverse event definitions significantly enhance zero-shot event detection performance on out-of-distribution data, underscoring their crucial role in event definition comprehension. 
On the other hand, the performance doesn't improve significantly with more than ten samples per event type. This is attributed to the reliance of zero-shot event detection on the model's ability to generalize to new event types and definitions. While a few samples aid in learning the meaning of event types, an excessive number is unnecessary.
In addition to scaling, we explore the impact of incorporating event ontology information in event definition and utilizing hard-negative samples during training. 
We observe that incorporating both components enhances the model's comprehension of event boundaries, resulting in higher recall and F1 scores.\looseness=-1


Following this finding, we further train our model on the DivED and Geneva~\cite{parekh2023geneva} datasets, and achieve state-of-the-art zero-shot event detection performance on ACE~\cite{doddington-etal-2004-automatic}, M2E2~\cite{li-etal-2020-cross} and MEE~\cite{veyseh2022mee} test sets benchmarked in TextEE~\cite{huang2023reevaluation}, surpassing strong LLMs such as Chat-GPT with less than 5 percents of model parameters, showing the effectiveness and efficiency of our method. \looseness=-1
To conclude, our main contributions are as follows:

\begin{enumerate}
  \item We design a data generation pipeline to generate a \textbf{Diverse Event Definition dataset (DivED)} with 3000+ event types and 10 diverse event definitions for each type. Our experiments reveal that diverse event types and event definitions are crucial to improve zero-shot ED.\looseness=-1
  \item We systematically study the \textbf{impact of various components of EE training data} on the ability of large instruction-tuned models to follow event definitions.
  \item Our \textbf{proposed model achieves SOTA results} on ACE, M2E2, and MEE datasets, surpassing GPT-3.5-Turbo model with drastically fewer parameters. \looseness=-1
\end{enumerate}




\section{Method}

\begin{figure*}[t]
\centering
\includegraphics[width=1.0\textwidth]{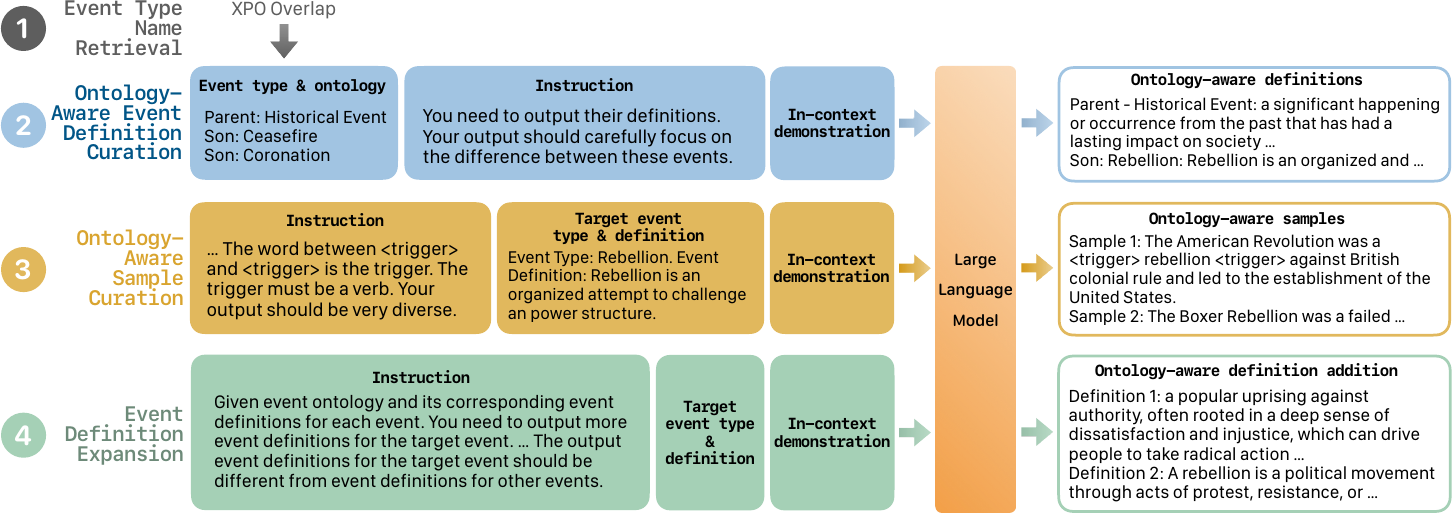}
\caption{Data generation pipeline to generate DivED dataset. The pipeline includes five main steps: (1) Event Type Name Retrieval: retrieve events from XPO overlap~\cite{spaulding-etal-2023-joint}; (2) Ontology-Aware Event Definition Curation: generate event type definitions for the event types retrieved from (1); (3) Ontology-Aware Sample Curation: generate samples for the retrieved event type names from (1) and event definition from (2); (4) Event Definition Expansion: Paraphrase and expand the event definition from (2), and (5) Ontology Pruning: Prune out events with high trigger overlap. 
Details of our prompt templates can be found in \autoref{sec:templates_for_data_generation}.}
\label{figure:data_pipeline}
\end{figure*}

In this section, we describe our data generation pipeline to generate the \textit{Diverse Event Definition Dataset (DivED}) and our systematic study on the impact of various components within event detection training data. We investigate how (1) the scaling of event types, event definition, and training samples and (2) incorporating ontology information and hard-negative samples can impact models' generalization to unseen event types.

\subsection{DivED Dataset Generation}
\label{subsection:data_generation_pipeline}
An event detection dataset with diverse event types and definitions is necessary to investigate the effect of training data components systematically.
Thus, we propose an automatic data generation pipeline that leverages proprietary LLM to generate \textit{Diverse Event Definition Dataset (DivED)}.
The data generation pipeline includes (1) Event Type Name Retrieval, (2) Ontology-Aware Event Definition Curation, (3) Ontology-Aware Sample Curation, (4) Event Definition Expansion, and (5) Ontology Pruning.
\autoref{figure:data_pipeline} illustrates the five-step data generation pipeline.
The examples of DivED and the data generation pipeline templates can be found in ~\autoref{sec:templates_for_data_generation}.

\paragraph{Step 1: Event Type Name Retrieval}
We follow ~\cite{zhan2023glen} methods to collect around 6000 event type names with ontology (dependency trees) from XPO-overlap~\citep{spaulding2023darpa}, which provides a large set of event entities that occurred in Wikidata.\footnote{\url{wikidata.org}}
To guarantee the testing events from ACE~\cite{doddington2004automatic}, M2E2~\cite{li2020cross}, and ~\cite{veyseh2022mee} datasets are held out for our later experiments, we manually filtered out all events that share the same dependency trees with these testing events.

\paragraph{Step 2: Ontology-Aware Event Definition Curation}\label{para:step_2}

After acquiring event type names and ontology (dependency trees of events), we instruct the model to simultaneously generate concise definitions for all event types within the ontology. Using one manually curated in-context example, we guide the model to differentiate similar events within the same ontology, resulting in distinct and well-distinguished event definitions, as demonstrated in \autoref{table:qualitative_examples}.

\begin{table*}[htbp]
    \small
    \centering
    \begin{tabularx}{16cm}{|p{1.1cm}|X|X|p{1.1cm}|}
        \hline
        Event & Event Definition & Sample & Trigger
        \\
        \hline
        Arriving
        &
        \textbf{Event Definition 1: } The act of Arriving involves the physical or virtual arrival at a destination 
        ...
        \textbf{Event Definition 10: } The Arrival event captures the moment when someone
        ...
        &
        \textbf{Sample 1: } The school field trip participants \textbf{arrived} at the museum and were greeted
        ...
        \textbf{Sample 10: } The visitors \textbf{arrived} at the aquarium and were led to the dolphin show by the staff.
        & arrived, ...
        \\
        
        \hline
        Drop in on
        &
        \textbf{Event Definition 1: } Drop in on refers to an unplanned and impromptu visit to a friend or acquaintance 
        ...
        \textbf{Event Definition 10: } The act of drop in on signifies an unscheduled visit to an individual's place
        ...
        &
        \textbf{Sample 1: } Renee decided to \textbf{pop in} on her friend who lived nearby and catch up.
        ...
        \textbf{Sample 10: } Jane had some free time on her hands and wanted to \textbf{pay a visit} to her former college roommate who lived close by.
        & pop in, pay a visit, ...
        \\

        \hline
        Visiting scenario arrival
        &
        \textbf{Event Definition 1: } Visiting scenario arrival entails arriving at a planned destination
        ...
        \textbf{Event Definition 10: } The event of Visiting scenario arrival involves arriving 
        ...
        &
        \textbf{Sample 1: } The investors \textbf{arrived} at the company's headquarters for their business presentation.
        ...
        \textbf{Sample 10: } The family \textbf{reached} the theme park with their pre-booked ride tickets.
        &
        arrived, reached, ...
        \\

        \hline
    \end{tabularx}
\caption{
We demonstrate the generated event definition and samples of a few sibling event types in the DivED dataset.
During data curation, we specifically prompt models to generate distinct event type definitions and samples for these similar event types to enhance the diversity of the generated data.
}
\label{table:qualitative_examples}
\end{table*}

\paragraph{Step 3: Ontology-Aware Sample Curation}
We follow a similar method as in \textbf{Ontology-Aware Event Definition Curation} to prompt the model with relative event types, event definition, and one manually curated in-context example to generate ten samples for multiple event types simultaneously. Each generated sample includes an input sentence and an output trigger of the corresponding event type.  
The generated samples can be seen in \autoref{table:qualitative_examples}.

\paragraph{Step 4: Event Definition Expansion}
To get multiple event definitions for each event type, we prompt the model to expand or paraphrase the event definition ten times with the provided event type name, event definition, event ontology, and one manually curated in-context example. 

\paragraph{Step 5: Ontology Pruning}
After generating data for all event types, we further prune out duplicate events within the same event ontology by identifying their output trigger overlap.
Specifically, for an event ontology tree $\{\mathbf{e}_{1}, \mathbf{e}_{2}, ...\} \in \mathbf{E}$ with multiple event types and ten samples per event, we calculate the output trigger overlap ratio between two event types $\mathbf{e}_{i}, \mathbf{e}_{j}$ where $i \neq j$. The trigger overlap is measured by exact string matching each of the ten triggers in $\mathbf{e}_{i}$ with the ten triggers in $\mathbf{e}_{j}$.
If the overlap ratio of output triggers exceeds a certain threshold (in our implementation, it is 0.5), we will consider one of the two events as duplicate and remove it from our dataset. This way, we can guarantee that the event types and output triggers of our dataset are diverse.\looseness=-1



\subsection{Data Impact Analysis}
With the generated \textbf{DivED} dataset, we systematically study the impact of various components in training data to understand how to instruction-tune the model with improved event definition following ability.\looseness=-1

\paragraph{Scaling of data components}
In \autoref{fig:teaser}, we show the data components within the training data. During training, we will provide several samples, each corresponding to an event type and definition, to query the models about the event trigger.
In testing time, we will further test on the unseen events, in which all the event types, definitions, and samples are unseen from training. This requires the model to generalize to the unseen events to be able to perform well during testing.
Following this intuition, we aim to investigate how different numbers of events, definitions, and samples can influence models' performance.
Specifically, for each dataset component, we fix the quantity of other components and evaluate the scaling law associated with it. For example, to investigate the scaling of event definition, we will use different number of event definitions per event, with a fixed amount of event type and samples.\looseness=-1

\paragraph{Ontology information}
We further look into the construction of event definition and negative samples.
In most zero-shot EE methods, they solely provide the information (event type, event definition) of the current event, without providing the event ontology information. We explore adding ontology information to the input definition in order to see how it helps models with the understanding of the event, and generalize better to unseen event types. The additional ontology information includes the parent and child events of an event ontology, as shown in \autoref{fig:teaser}.

\paragraph{Hard-negative samples}
During model training, we use input sentences paired with output triggers. Positive samples are based on ground truth events, where the output for event trigger identification is not "None." To incorporate predictions for "None" events, we create negative samples by prompting the model with input sentences and an event type that does not occur in the sentence.
In this work, we aim to explore how integrating ontology information into negative sample construction affects the model's ability to learn event definitions and boundaries. Instead of randomly assigning unrelated events during negative sample creation, we will assign sibling event types to form \textbf{hard-negative samples}. These challenging examples may offer additional signals about event boundaries that aid the model in distinguishing between similar events and improve its understanding of event definitions.

\section{Experimental Setup}
\begin{figure*}[t!]
    \centering
    \includegraphics[width=\textwidth]{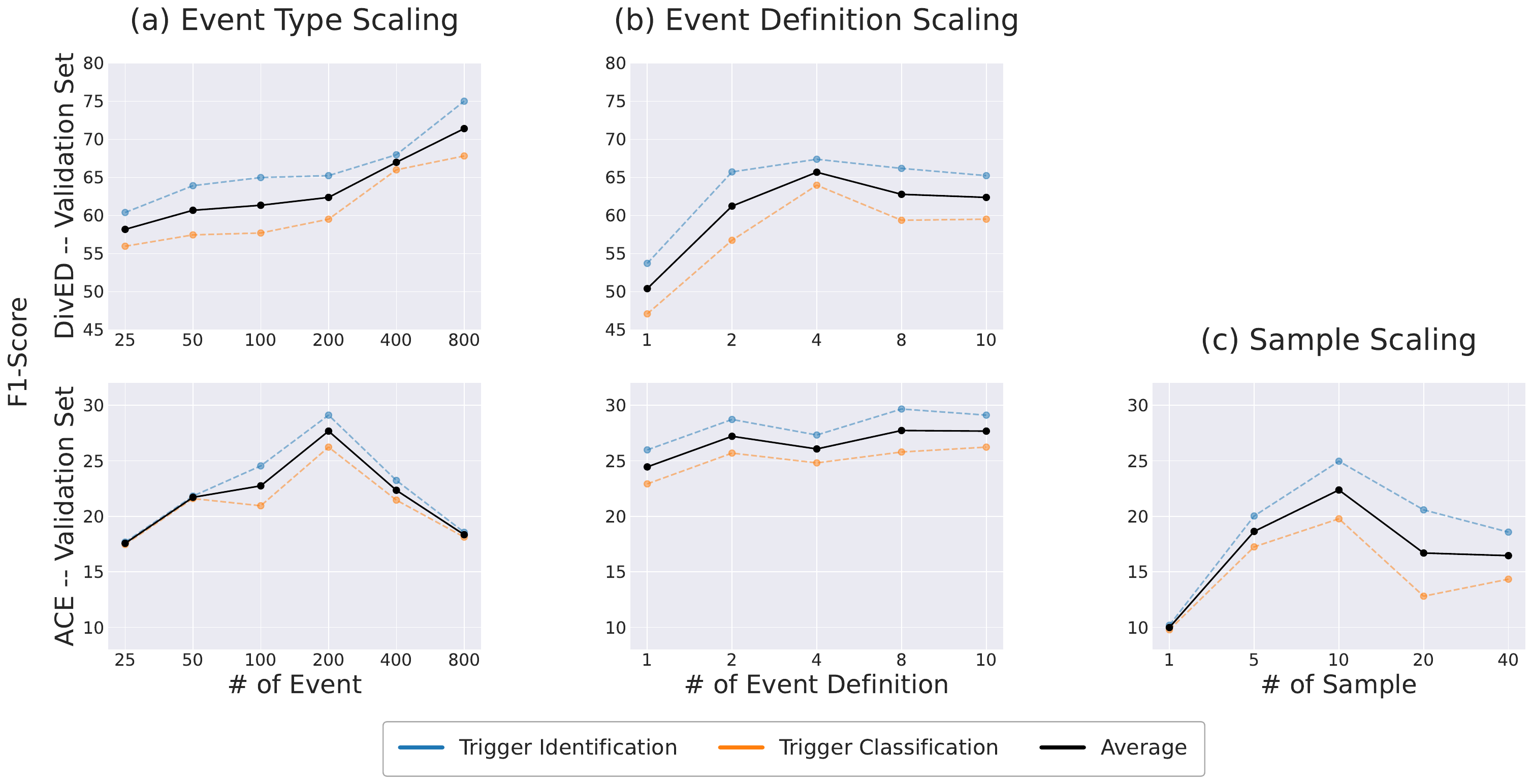}
    \vspace{-1em}
    \caption{
    The scaling of different dataset components. We train the models with different number of event types, event definitions per event type and samples per event type. After training, we further report the F1 scores on \textit{DivED -- Validation} and  \textit{ACE Validation set}. Note that we do not report the \textit{DivED -- Validation} score separately for sample scaling as we utilize the Geneva~\cite{parekh2023geneva} train set to explore sample scaling rather than DivED train set. \looseness=-1
    }
    \vspace{-0.5em}
    \label{fig:scaling-law}
\end{figure*}
In this section, we first describe the details of \textbf{Data Impact Analysis} experiments, which analyze the impact of different data components. 
We further describe baselines and training details in \textbf{Enhancing Zero-Shot Event Detection}, in which we integrate the optimal settings from Data Impact Analysis to train our zero-shot event detection model and compare to previous state-of-the-art large language models (LLMs) on three event extraction datasets.


\subsection{Data Impact Analysis}
\paragraph{Model and Training Details}
We utilize LLaMA-2-7B~\cite{touvron2023llama} for our experiments to examine the impact of various data components. Employing a batch size of 96, a learning rate of 2e-5, and training for 20 epochs. We consistently use the DivED dataset with unified variables: 200 Event Types, 10 Event Definitions, 10 Samples, and 10 negative samples per sample. In each scaling experiment, only one variable is scaled while others are fixed. Notably, ACE-related events are excluded from the DivED dataset to ensure the test events are entirely novel. 
\paragraph{Event type and event definition scaling}
We experiment on [200, 400, 800, 1600, 3200] events on the DivED dataset for event-type scaling. For Event definition scaling, we test [1, 2, 4, 8, 10] event definition to investigate the scaling law of these variables.
\paragraph{Sample scaling}
For sample scaling, since DivED only has ten samples per event, we conduct it on the Geneva~\cite{parekh2023geneva} dataset and test with [1, 5, 10, 20, 40] samples per event. We filter out all ACE-related events for the Geneva dataset to make sure the test events are unseen.
\paragraph{Event Ontology and Hard Negative Samples}
In event ontology experiments, we assess two settings: with or without event ontology. For hard negative samples, we experiment using zero or three hard negative samples within the ten negative samples. Evaluation is conducted on the ACE dataset for both experiments.
\paragraph{Evaluation}
We report the F1 scores of \textbf{event trigger identification} and \textbf{event trigger classification} on DivED (in-domain) and ACE (out-domain) validation set. In in-domain evaluation, 50 unseen events from DivED (absent in training) are tested. For out-domain assessment, the model is tested on the ACE dataset, comprising 33 event types unseen during training. We analyze the models' zero-shot generalization on these sets, presenting a comparison in \autoref{fig:scaling-law}.
Notably, DivED (in-domain) exhibits similar event definition and sample length to the training set, while ACE (out-domain) has longer definitions and a different writing style, focusing on argument details alongside the event itself.

\begin{table}[t]
\centering
\small
\setlength{\tabcolsep}{1mm}
\begin{NiceTabular}{p{0.46\textwidth}}
\Hline
\multicolumn{1}{c}{DivED-Train -- Event: Money Laundering}\\
\hline
\textbf{Event Definition}:\\
\textit{Money Laundering is the process of moving illicit funds through complex financial transactions, disguising the origin of the money and making them seem legitimate.}\\
\textbf{Avg. Def. Length}: 42.9; \quad \textbf{Avg. Sample Length}: 23.3 \\
\Hline
\Hline
\multicolumn{1}{c}{DivED-Validation -- Event: Ceasefire}\\
\hline
\textbf{Event Definition}:\\
\textit{A ceasefire is a mutual agreement between opposing armed groups to halt all aggressive actions and refrain from initiating any new hostilities, often negotiated to allow for the delivery of aid and the creation of safe zones for civilians.}\\
\textbf{Avg. Def. Length}: 42.2; \quad \textbf{Avg. Sample Length}: 23.3 \\
\Hline
\Hline
\multicolumn{1}{c}{Ace-Validation -- Event: BE-BORN}\\
\hline
\textbf{Event Definition}:\\
\textit{BE-BORN Event occurs whenever a PERSON Entity is given birth to. Please note that we do not include the birth of other things or ideas.}\\
\textbf{Avg. Def. Length}: 65.3; \quad \textbf{Avg. Sample Length}: 35.6 \\
\Hline

\end{NiceTabular}
\caption{
Dataset comparison. We show the comparison of the definition, average definition token length and average sample token length between \textbf{DivED train set}, \textbf{DivED validation set} and \textbf{ACE test set}.}
\vspace{-0.5em}
\label{table:compare_data}
\end{table}

\subsection{Enhancing Zero-Shot Event Detection}
\paragraph{Training Details}
Following our observation in \textbf{Data Impact Analysis}, we further train the LLaMA-2-7B~\cite{touvron2023llama} models on DivED and Geneva~\cite{parekh2023geneva} dataset following the optimal setting. We use 200 + 90 event types from DivED and Geneva~\cite{parekh2023geneva} datasets. We use ten event definitions, ten samples, and ten negative samples per sample for each event type while incorporating the ontology information and three hard-negative samples.
\paragraph{Baselines}
\label{section:baselines}
In our experiments, we conduct a comparison between our finetuned LLaMA-2-7B~\cite{touvron2023llama} and several zero-shot event detection baselines, including ChatGPT~\cite{chatgpt_cite}, ChatIE \cite{wei2023zero} and \cite{gao2023exploring} and LLaMA-2-Geneva. Prompt templates for the baselines are provided at Appendix \ref{sec:templates_for_experiments}.
\begin{itemize}
    \item \textbf{ChatGPT~\cite{chatgpt_cite}}: GPT-3.5-Turbo was prompted with the proposed method for a fair comparison with finetuned LLaMA-2-7B.
    \item \textbf{ChatIE \cite{wei2023zero}}: ChatIE is a framework that transforms the zero-shot event detection task into a multi-turn question-answering problem. Here, LLMs are first prompted (as shown in Appendix \ref{sec:templates_for_experiments}) to identify the event type and then sequentially prompted to identify the trigger.  
    \item \textbf{\citet{gao2023exploring}}: This work explores the feasibility of ChatGPT as a zero-shot event detection model and further analyses the impact of event definitions, in-context examples and counterfactual examples in the prompt template presented at Appendix \ref{sec:templates_for_experiments}. We prompt ChatGPT with event definitions and positive examples in our implementation as this setup performed best on \citet{gao2023exploring} evaluation.\looseness=-1
    \item \textbf{LLaMA-2-Geneva}: We additionally train an LLaMA2-7B model~\cite{touvron2023llama} on Geneva~\cite{parekh2023geneva} datasets as a transfer learning baseline. We first filter out all events related to ACE, M2E2, and MEE datasets from the training set, leaving 90 event types. We further train the model with all samples on the remaining event types. 
\end{itemize}

\paragraph{Evaluation Datasets}
\label{section:datasets}
Our experiments comprehensively compare our fine-tuned LLaMa-2-7B with baselines across three popular event extraction benchmarks, including ACE05, M2E2, and MEE. We consider the English annotations of these datasets and report the F1 scores of event trigger identification and event trigger classification on their test set process by TextEE~\cite{huang2023reevaluation}.
\begin{itemize}
    \item ACE05~\citep{doddington2004automatic} is an end-to-end event extraction dataset which covers texts from several sources such as newswire, broadcast news and weblogs. 
    \item M2E2~\citep{li2020cross} is an end-to-end event extraction dataset collected from the multimedia domain. We only consider the text part.
    \item MEE~\citep{veyseh2022mee} is a multilingual end-to-end event extraction dataset collected from Wikipedia which is extended from MINION~\citep{song2015light}.
\end{itemize}

\section{Results}
\begin{table}[t]
\centering
\small
\setlength{\tabcolsep}{0.3mm}
\begin{NiceTabular}{l||ccc|ccc}
\Hline
\Block{1-1}{Metric $\rightarrow$\\Model $\downarrow$}
&\Block{1-1}{\\Prec.} & \Block{1-1}{Trigger ID\\Rec.} & \Block{1-1}{\\F1}
&\Block{1-1}{\\Prec.} & \Block{1-1}{Trigger CLS\\Rec.} & \Block{1-1}{\\F1}
\\
\Hline
Ours & 45.3 & 22.0 & 29.1 & 36.6 & 20.4 & 26.2 \\
w/o Ontology & 55.0 & 11.8 & 19.4 & 34.3 & 11.1 & 16.7 \\
w/o Hard Neg. & 53.5 & 16.6 & 25.3 & 45.5 & 15.6 & 23.2 \\
\Hline
\end{NiceTabular}
\caption{
We report the experiment results of providing ontology information and using hard negative samples.
}
\vspace{-1em}
\label{table:ontology-hardneg}
\end{table}

\begin{table*}[t]
\centering
\small
\setlength{\tabcolsep}{0.3mm}
\begin{tabular}{l|ccc|ccc||ccc|ccc||ccc|ccc||c|c}
\hline
\multicolumn{1}{l}{}  & \multicolumn{6}{c}{\bf ACE} & \multicolumn{6}{c}{\bf M2E2} & \multicolumn{6}{c}{\bf MEE} & \multicolumn{2}{c}{\bf Average}  \\
\hline
\multicolumn{1}{l|}{Metric $\rightarrow$} & \multicolumn{3}{c|}{Trig. ID} & \multicolumn{3}{c||}{Trig. CLS} & \multicolumn{3}{c|}{Trig. ID} & \multicolumn{3}{c||}{Trig. CLS} & \multicolumn{3}{c|}{Trig. ID} & \multicolumn{3}{c||}{Trig. CLS} & Trig. ID & Trig. CLS \\ 
Model $\downarrow$ & Prec. & Rec. & F1 & Prec. & Rec. & F1 & Prec. & Rec. & F1 & Prec. & Rec. & F1 & Prec. & Rec. & F1 & Prec. & Rec. & F1 & F1  & F1 \\ 
\hline
ChatIE &

4.8 &
11.9 &
6.8 &

2.5 &
6.4 &
3.6 &

4.3 &
31.4 &
7.5&

2.6 &
19.2 &
4.6 & 

12.4 &
31.2 &
17.2 &

7.5 &
24.2 & 
11.9 & 10.5 & 6.7
\\

\citeauthor{gao2023exploring} &
42.5 &
28.7 &
34.2 &

30.6 &
20.6 &
24.6 &

16.6 &
38.0  &
23.1 &

14.5 &
33.2 &
20.2 &

84.7 & 
7.8 & 
14.2 &

77.9 &  
7.15 & 
13.9 &

23.8 & 19.6

\\

GPT-3.5 & 

9.2 &
60.7 &
15.9 &

3.6 &
43.9 &
6.6 &

7.0 &
63.3 &
12.6 &

4.9 &
54.2 &
8.9 &

12.5 &
33.2 & 
18.2 &

7.6 & 
24.4 & 
11.6 & 15.6 & 9.0

\\


Geneva &
47.7 &
14.5 &
22.2 &

18.2 &
13.8 &
15.7 &

19.1 &
17.9 &
18.5 &

17.9 &
17.0 &
17.4 &

70.5 &
24.5 &
\textbf{36.4} &

63.0 &
24.1 &
\textbf{34.9} & 25.7 & 22.6\\

\hline
Ours & 
46.7 &
26.9 &
\textbf{34.2} &

36.7 &
24.1 &
\textbf{29.1} &

21.2 &
26.1 &
\textbf{23.4} &

19.8 &
24.7 &
\textbf{22.0} &

70.9 &
16.7 &
27.1 &

65.7 &
16.2 &
26.0 & \textbf{28.2} & \textbf{25.7}\\

\hline
\end{tabular}
\caption{
The experiment results on ACE ,M2E2 and MEE test set. We compare the performance of LLaMA-2-7b training on DivED dataset (Ours) with ChatIE \citep{wei2023zero}, \citet{gao2023exploring}, GPT3.5 model and LLaMA-2-7b trained on Geneva \citep{parekh2023geneva} dataset.  We report the Precision, Recall and F1 scores. We also report the average F1 score across all datasets. We abbreviate Trigger as Trig.\looseness=-1}
\label{table:main-table}
\end{table*}

\subsection{Data Impact Analysis}

\paragraph{Scaling of event types}
In \autoref{fig:scaling-law} column (a), we show the results of training a LLaMA-2-7B model with different numbers of events. It is seen that scaling up the number of events consistently helps the model performance on the in-domain DivED validation set. However, while training the model with more events continuously scales up the performance on the ACE validation set under 200 events, using more than 200 events leads to degeneration of the performance. This can be caused by the model overfitting to the domain of training data. While we continuously train on new events, the model can still overfit to the domain of the data itself, for example, the format of the event definition and samples. This also shows that while our generated \textbf{DivED} dataset has a large number of events, the generated samples and event definition might still have spurious correlations that can lead to overfitting.\looseness=-1

\paragraph{Scaling of event definition}
In \autoref{fig:scaling-law} column (b), we show the results of event definition scaling. On both the DivED and ACE validation sets, the performance scales up with four event definitions per event. While using more than four event definitions does not help the in-domain performance, it can help the model generalize better to the out-domain test set. This shows that adding more diverse event definitions during training can further improve the model's robustness. Helping the model to generalize to more diverse event formats during inference, thus improving zero-shot performance.\looseness=-1

\paragraph{Scaling of samples}
To evaluate how using more samples helps the model's zero-shot generalization, we experiment using the Geneva dataset, which has a large number of high-quality samples per event type, and test on the ACE validation set. The results are shown in \autoref{fig:scaling-law} column (c). Surprisingly, while using more samples usually helps models' performance in a supervised setting, using more than ten samples hurts models' performance. This can be caused by the model overfitting the training data and becoming less robust to unseen events.\looseness=-1

Following the results above, we conclude that the key to improving models' zero-shot generalization to unseen events is to use a diverse set of event definitions with a certain amount of event types and samples. While a small amount of event type and samples helps, using too much can make the models overfit to the training source, leading to a degeneration of the generalization ability. This effect can be specifically obvious in machine-generated data, which can have spurious correlations and lack diversity in certain aspects. \looseness=-1

\paragraph{Event ontology and hard-negative samples}
In \autoref{table:ontology-hardneg}, we further investigate the usefulness of the event ontology and hard-negative samples. It can be seen that after removing the event ontology or hard-negative samples, fewer triggers are predicted, which leads to a much lower recall and F1 score. This means that the model becomes more conservative at predicting triggers. We hypothesize that the model can be trained to distinguish similar events by providing ontology and hard negative samples. At testing time, this can help the models be more certain at predicting the triggers and increase the number of matching triggers.\looseness=-1

\subsection{Enhancing Zero-Shot Event Detection}
Following the observation from \textbf{Data Impact Analysis}, we further apply the best setting and compare it with baseline models described in \autoref{section:baselines}. We show the results in \autoref{table:main-table}.
It can be seen that LLaMA-2-7B trained on the DivED dataset (Ours) consistently outperforms GPT baselines (ChatIE, \citet{gao2023exploring} and GPT-3.5) on all ACE, M2E2, and MEE datasets and surpasses our LLaMA2-Geneva baselines on ACE and M2E2 datasets. 
For the MEE dataset, LLaMA2-Geneva achieves the best performance. Upon further investigation into the predicted results, we found that LLaMA2-Geneva can better predict samples that have multiple ground truth event triggers in one event type, which frequently occurred in MEE Geneva datasets but less occurred in ACE, M2E2 and DivED datasets, directly leading to higher Recall and F1 scores on MEE dataset.
Generally, our proposed model achieves the best average F1 scores on both Event Trigger Identification and Event Trigger Classification, showing the superiority of the training method.

\section{Discussion}

\begin{table}[t]
\centering
\small
\setlength{\tabcolsep}{1mm}
\begin{NiceTabular}{l||c|c|c|c}
\Hline
\Block{1-1}{Metric $\rightarrow$\\Model $\downarrow$} 
& \Block{1-1}{Trigger ID \\ F1} & $\Delta$ & \Block{1-1}{Trigger CLS \\ F1} & $\Delta$ \\
\Hline
Ours & 29.1 & & 26.2 &\\
w/o Def & 14.15 & -52\% & 10.83 & -59\% \\
\Hline
Geneva & 25.2 & & 17.4 &\\
w/o Def & 23.44 & -7\% & 8.2 & -53\% \\
\Hline
\end{NiceTabular}
\caption{
We assess model performance drop by removing event definitions during training. We compare LLaMA-2 models trained on Geneva and DivED datasets. A higher drop rate indicates greater reliance on event definitions. \looseness=-1}
\vspace{-1em}
\label{table:follow-def}
\end{table}

\subsection{Do Models Follow the Event Definition?}

Instruction-tuned models excel in various zero-shot tasks but can excessively rely on the spurious patterns within the provided prompt, neglecting instruction semantics~\cite{kung-peng-2023-models, yin2023did}. 
In this work, we aim to enhance zero-shot event detection by training model with better event definition following. 
To assess the model's event definition following ability, we conduct an ablation study following prior work's setting~\cite{kung-peng-2023-models}, comparing our LLaMA-2-7B model trained on DivED data with one trained on the conventional EE dataset Geneva~\cite{parekh-etal-2023-geneva}. 
To verify whether our proposed model have better event definition following ability compare to models learning on convention EE datasets, we follow prior work ~\cite{kung-peng-2023-models} to conduct an ablation study. We compare our LLaMA-2-7B~\cite{touvron2023llama} model trained on DivED data with a LLaMA-2-7B model trained on conventional EE dataset such as Geneva~\cite{parekh-etal-2023-geneva}. Despite having numerous samples per event, conventional EE dataset has only one definition per event type, which largely differs from DivED dataset. 
We report the performance drop rate after removing the event definition during training and testing for both models in Table \autoref{table:follow-def}. It can be seen that while the performance drops for both models after removing the event definition during training, the model trained on the DivED dataset has a higher performance drop, especially in \textit{Event Trigger Identification}, showing that our proposed model heavily relies on event definition during training and inference. This indicates that our model is better at utilizing the event definition information, potentially exhibiting a better event definition following ability.\looseness=-1
\section{Related Work}

\paragraph{Low-resource information extraction}
Low-resource IE models secure their performance with limited training data by cross-task transfer learning that uses supervision from tasks like Semantic Role Labeling~\cite{Zhang2021ZeroshotLabelAwareEvent,Huang2018ZeroShotTransferLearning}, indirect supervision that reformulates the task as data-rich tasks like NLI or QA~\cite{Xu2023CanNLIProvidea,Sainz2022TextualEntailmentEvent,Ma2023ParameterEfficientLowResourceDialogue,lu-etal-2022-summarization}, both heavily rely on task compatibility. Some works focus on prompting generative LMs with enriched task requirements and examples~\cite{Li2023EvaluatingChatGPTInformation,Gao2023ExploringFeasibilityChatGPT}, which is constrained the diversity of human-curated training data. In this work, we tackle the zero-shot IE task by expanding the diversity and scope of the seed data set with LLM without the need for cross-task resources or human annotation.

\paragraph{Data generation for IE}
Existing works explore different strategies to generate training data instances given a known task output space through analogous input~\cite{Kumar2020DataAugmentationUsing,lee2021neural}, creating pesudo labels with weak annotator~\cite{He2021GenerateAnnotateLearn,Chia2022RelationPromptLeveragingPromptsa,Ye2022ZeroGenEfficientZeroshot,Schick2021GeneratingDatasetsPretrained}, reverse generation~\cite{Meng2022GeneratingTrainingData,Gao2021MakingPretrainedLanguage,Josifoski2023ExploitingAsymmetrySynthetic} and structure-to-text generation~\cite{ma2023star}. Different from introducing more data instances for observed task space, we instead aim to extend the model's generalizability by generating new types and their definitions for unseen data distribution that extend the task space with LLM-oriented data generation.
\section{Conclusion}
We investigate how incorporating diverse event types and definitions benefits zero-shot event detection models. The proposed DivED dataset features a large number of diverse event types and definitions, which helps train the model to better generalize to unseen event definitions.
By further incorporating event ontology and hard negative samples, we finetuned a LLaMA-2 model on DivED and Geneva datasets, which consistently surpasses previous SOTA ChatGPT prompting baselines in zero-shot ED on ACE, M2E2, and MEE datasets. Overall, our findings provide insights to improve models' event definition following ability and provide an opportunity to further advance zero-shot ED on open-sourced models.

\section*{Limitation}
Our study on zero-shot event detection, despite its advances, faces several limitations. The reliance on automatically generated datasets may not fully capture complex real-world events, potentially limiting the model's generalizability. Additionally, the effectiveness of our approach depends on detailed event ontology and the availability of hard-negative samples, which might not always be accessible. Scalability also poses a challenge, as expanding the diversity of event types requires significant computational and data resources. Moreover, our findings are primarily based on English language benchmarks, raising questions about the applicability of our results across different languages and domains. Future research should address these limitations to enhance the robustness and universality of zero-shot event detection models.

\bibliography{anthology,custom,ma_auto}

\newpage
\appendix

\section{Cost Estimates for OpenAI}
\label{sec:gpt4 cost}

We implement all baselines in Section \ref{section:baselines} with GPT-3.5-Turbo. The estimated cost of implementing our baselines is approximately 100 USD.  Similarly, the estimated cost of implementing our baselines on GPT-4 will be approximately 3000 USD and we leave this implementation for future due to limited resources. This further emphasizes that our finetuned model surpasses larger LLMs in performance as well as accessibility due to the cost effectiveness of the method. 

\section{Data Generation of DivED dataset}
\label{sec:templates_for_data_generation}


\paragraph{Step 1: Event Type Name Retrieval}
We follow ~\cite{zhan2023glen} methods to collect around 6000 event type names with ontology (dependency trees) from XPO-overlap~\citep{spaulding2023darpa}, which provides a large set of event entities that occurred in Wikidata.\footnote{\url{wikidata.org}}
To guarantee the testing events from ACE~\cite{doddington2004automatic}, M2E2~\cite{li2020cross}, and ~\cite{veyseh2022mee} datasets are held out for our later experiments, we manually filtered out all events that share the same dependency trees with these testing events.

\paragraph{Step 2: Ontology-Aware Event Definition Curation}\label{para:step_2}

After acquiring event type names and ontology (dependency trees of events), we instruct the model to simultaneously generate concise definitions for all event types within the ontology. Using one manually curated in-context example, we guide the model to differentiate similar events within the same ontology, resulting in distinct and well-distinguished event definitions, as demonstrated in \autoref{table:qualitative_examples}.
The template utilized in this step is presented at Table~\ref{table:template_definition_generation}.

\paragraph{Step 3: Ontology-Aware Sample Curation}
We follow a similar method as in \textbf{Ontology-Aware Event Definition Curation} to prompt the model with relative event types, event definition, and one manually curated in-context example to generate ten samples for multiple event types simultaneously. Each generated sample includes an input sentence and an output trigger of the corresponding event type.  
The generated samples can be seen in \autoref{table:qualitative_examples}.
The template utilized in this step is presented at Table~\ref{table:template_sample_generation}.

\paragraph{Step 4: Event Definition Expansion}
To get multiple event definitions for each event type, we prompt the model to expand or paraphrase the event definition ten times with the provided event type name, event definition, event ontology, and one manually curated in-context example. 
The template utilized in this step is presented at Table~\ref{table:template_definition_rewrite}.

\paragraph{Step 5: Ontology Pruning}
After generating data for all event types, we further prune out duplicate events within the same event ontology by identifying their output trigger overlap.
Specifically, for an event ontology tree $\{\mathbf{e}_{1}, \mathbf{e}_{2}, ...\} \in \mathbf{E}$ with multiple event types and ten samples per event, we calculate the output trigger overlap ratio between two event types $\mathbf{e}_{i}, \mathbf{e}_{j}$ where $i \neq j$. The trigger overlap is measured by exact string matching each of the ten triggers in $\mathbf{e}_{i}$ with the ten triggers in $\mathbf{e}_{j}$.
If the overlap ratio of output triggers exceeds a certain threshold (in our implementation, it is 0.5), we will consider one of the two events as duplicate and remove it from our dataset. This way, we can guarantee that the event types and output triggers of our dataset are diverse.\looseness=-1

\paragraph{Generated Samples}
\autoref{table:example_GenData_1}, \autoref{table:example_GenData_2} and \autoref{table:qualitative_examples_full} present examples of DivED dataset.

\begin{table*}
\centering
\begin{tabular}{p{0.98\linewidth}}
\toprule
\textbf{Instruction:}\\
Now I will give you some ontology, and then you need to output their definitions. Your output should carefully focus on the difference between these events.\\
\midrule
\textbf{In-Context Demonstration:}\\
Input: \\
parent event: ...\\
son event: ...\\
\\
Output:\\
parent event: event definition of parent event ... \\
son event: event definition of son event ... \\
\midrule
\textbf{Instruction:}\\
Now I will give you some event ontology, and your output should carefully follow the previous examples. Your output should carefully focus on the event ontology.\\
\midrule
\textbf{Query:}\\
Input:\\
parent event: mental state\\
son event:  belief\\
son event:  doubt\\
son event:  awakening\\
son event:  emotional state\\
son event:  intention\\
son event:  mental confusion\\
son event:  understanding\\
\midrule
\textbf{GPT Output:} \\
Mental State: A mental state refers to an individual's current psychological condition or state of mind. It encompasses various emotions, thoughts, and cognitive processes that a person may experience at any given time.\\
Belief: Belief is a mental state characterized by the acceptance or conviction that something is true or exists, often without concrete evidence. It involves having confidence in the validity of a particular idea, concept, or proposition.\\
Doubt: Doubt is a mental state characterized by uncertainty or skepticism regarding the truth or validity of a belief or idea. It involves questioning or being unsure about the accuracy of a particular belief or assertion.\\
Awakening: Awakening refers to a profound shift in an individual's perception or understanding. It is often associated with a realization or enlightenment, leading to a significant change in one's perspective or awareness, often of a spiritual or philosophical nature.\\
Emotional State: An emotional state relates to a person's feelings and emotions at a specific moment. These states can include happiness, sadness, anger, fear, or any other emotional response to internal or external stimuli.\\
Intention: Intention is a mental state involving a purposeful and deliberate desire to perform a particular action or achieve a specific goal. It is often accompanied by a plan or strategy to accomplish the intended outcome.\\
Mental Confusion: Mental confusion is a mental state characterized by a lack of clarity or cognitive disarray. It involves difficulty in processing information, making decisions, or understanding one's surroundings, leading to a sense of disorientation.\\
Understanding: Understanding is a mental state involving comprehension and insight into a concept, idea, or situation. It signifies the ability to grasp and make sense of information or knowledge, often leading to increased knowledge or awareness.\\
\bottomrule
\end{tabular}
\caption{Template for Ontology-Aware Event Definition Curation}
\label{table:template_definition_generation}
\end{table*}

\begin{table*}
\centering
\begin{tabular}{p{0.98\linewidth}}
\toprule
\textbf{Instruction:}\\
Here we provide the definition and examples of three event types. The word between <trigger> and <trigger> is the trigger. The trigger must be a verb. Your output should be very diverse.\\
\midrule
\textbf{In-Context Demonstration:}\\
Input: \\
Event type: ...\\
Event Definition: ...\\
\\
Output:\\
Sample 1: ... \\
Sample 2: ... \\
\midrule
\textbf{Instruction:}\\
Please generating a new event type following the same format. Note that there is an ontology, we provide event types and their definitions.\\

\midrule
\textbf{Instruction:}\\
Please generate samples for event rebellion.
Your output should carefully consider the ontology. Your output examples should not be examples for other son events. \\

\midrule
\textbf{Query:}\\
Event Ontology:\\
parent event: historical event\\
son event:  ceasefire\\
son event:  cornonation\\
son event:  rebellion\\
son event:  revolt\\

Event Type: rebellion \\
Event Definition: Rebellion is an organized and sustained attempt to challenge an established power structure, often involving large numbers of people and various forms of resistance, including civil disobedience and armed struggle. \\

\midrule
\textbf{GPT Output:} \\
Sample 1: The American Revolution was a <trigger>rebellion<trigger> against British colonial rule and led to the establishment of the United States. \\
Sample 2: The Syrian Civil War began as a <trigger>rebellion<trigger> against the government of President Bashar al-Assad. \\
Sample 3: The Boxer Rebellion was a failed <trigger>rebellion<trigger> against foreign influence in China in 1900. \\
Sample 4: The Haitian Revolution was a successful <trigger>rebellion<trigger> led by enslaved people against French colonial rule. \\
Sample 5: The Glorious Revolution of 1688 was a peaceful <trigger>rebellion<trigger> in England that resulted in the overthrow of King James II and the establishment of a constitutional monarchy. \\
Sample 6: The Arab Spring uprisings were a series of <trigger>rebellions<trigger> across the Middle East and North Africa in 2011 that aimed to overthrow authoritarian regimes. \\
Sample 7: The Boston Tea Party was an act of <trigger>rebellion<trigger> by American colonists in 1773 against British taxes on tea. \\
Sample 8: The Indian Rebellion of 1857 was a failed <trigger>rebellion<trigger> against British colonial rule in India. \\
\bottomrule
\end{tabular}
\caption{Template for Ontology-Aware Sample Curation}
\label{table:template_sample_generation}
\end{table*}

\begin{table*}
\centering
\begin{tabular}{p{0.98\linewidth}}
\toprule
\textbf{Instruction:}\\
Now I will give you an event ontology and its corresponding event definitions for each event.
You need to output more event definitions for the target event.
Your output should carefully focus on the difference between these events in the event ontology. The output event definitions for the target event should be different from event definitions for other events. And you need to try to give me diverse event definitions for the target event, but maintain the original semantics.\\
\midrule
\textbf{In-Context Demonstration:}\\
Input: \\
parent event: ...\\
son event: ...\\
\\
Output:\\
1. ... \\
\midrule
\textbf{Instruction:}\\
Now I will give you some event ontology, and your output should carefully follow the previous examples. Your output should carefully focus on the event ontology.\\
\midrule
\textbf{Query:}\\
Input:\\
parent event:  Mental State: A mental state refers to an individual's current psychological condition or state of mind. It encompasses various emotions, thoughts, and cognitive processes that a person may experience at any given time.\\
son event:  Belief: Belief is a mental state characterized by the acceptance or conviction that something is true or exists, often without concrete evidence. It involves having confidence in the validity of a particular idea, concept, or proposition.\\
son event:  Doubt: Doubt is a mental state characterized by uncertainty or skepticism regarding the truth or validity of a belief or idea. It involves questioning or being unsure about the accuracy of a particular belief or assertion.\\
son event:  Awakening: Awakening refers to a profound shift in an individual's perception or understanding. It is often associated with a realization or enlightenment, leading to a significant change in one's perspective or awareness, often of a spiritual or philosophical nature.\\

\midrule
\textbf{Instruction:} \\

Now please give me 10 more definitions for the event Awakening:
Your output event definitions should be diverse, but maintain the semantics of the original definition. 
Your output should follow the below format.\\

\midrule
\textbf{GPT Output:} \\
1. Awakening is a transformative experience marked by a deep and sudden insight that fundamentally alters a person's way of thinking or viewing the world, frequently triggering a newfound sense of purpose or clarity.\\
2. An Awakening signifies a profound shift in an individual's consciousness, usually triggered by a moment of profound understanding or revelation, which can lead to a complete reevaluation of their beliefs and values.\\
3. The concept of Awakening encompasses a powerful, introspective moment where an individual experiences a heightened awareness of their existence, often bringing about a renewed sense of meaning and direction in life.\\
...\\

\bottomrule
\end{tabular}
\caption{Template for ontology-aware event definition addition}
\label{table:template_definition_rewrite}
\end{table*}

\begin{table*}
\centering
\begin{tabular}{p{0.98\linewidth}}
\toprule
\textbf{Instruction:}\\
This is an event extraction task where the goal is to extract structured events from the text. A structured event contains an event trigger word and an event type.\\
\midrule
\textbf{Event Types and their definitions:} \\
Event type 1 : Event Definition 1\\
Event type 2 : Event Definition 2\\
:\\
:
\\ \\

\midrule
\textbf{In-Context Examples from the dataset} \\
\textbf{Positive Example 1}\\
Sentence 1: ... \\
Output: Trigger, Event Type\\

\textbf{Positive Example 2}\\
Sentence 2: ... \\
Output: Trigger, Event Type\\

\textbf{Positive Example 3}\\
Sentence 3: ... \\
Output: Trigger, Event Type \\
\\
\midrule
\textbf{Example 4}\\
Sentence : User Query \\
Output: \\
\bottomrule
\end{tabular}
\caption{Template for Guo et al }
\label{table:template_guo_baseline}
\end{table*}

\begin{table*}
\centering
\begin{tabular}{p{0.98\linewidth}}
\toprule
\textbf{Instruction:}\\
The list of event types: <list all event types for the dataset> \\
Give a sentence: <user query>. \\
What types of events are included in this sentence? Please return the most likely answer according to the list of event types above. Require the answer in the form: Event type.
\\
\midrule
\textbf{GPT Response:} \\
Event Type
\\ 

\midrule
\textbf{Instruction:} \\
If the event type is identified, return the event trigger word(s). Return 'NONE' if the event type is absent. Separate multiple event trigger words with semicolon (;).  Refrain from explaining your reasoning—provide only the direct answer. \\
Trigger - \\
\bottomrule
\end{tabular}
\caption{Template for multi-turn ChatIE}
\label{table:template_chatie_baseline}
\end{table*}

\begin{table*}
\centering
\begin{tabular}{p{0.98\linewidth}}
\toprule
\textbf{Instruction:}\\
Act as an AI assistant specialized in extracting events. When given a sentence, a specified event type, and its definition, examine the sentence for the event type. If the event type is identified, return the event trigger(s). Return 'NONE' if the event type is absent. Separate multiple event triggers with semicolon (;).  Refrain from explaining your reasoning—provide only the direct answer. \\
\\
Sentence: <Sentence>\\

Event Type: <Type of Event> \\

Event Definition: <Definition of the Event> \\
\\

TRIGGER:\\

\bottomrule
\end{tabular}
\caption{Template for GPT-3.5-Turbo. We prompt the model with the definition of each possible event type from the dataset and aggregate the predictions for evaluation. }
\label{table:template_chatgpt_baseline}
\end{table*}

\section{Templates for Experiments}
\label{sec:templates_for_experiments}

The compared baselines include ChatGPT, ChatIE \cite{wei2023zero} and \cite{gao2023exploring}. 
\begin{itemize}
    \item \textbf{ChatGPT}: ChatGPT were prompted with the proposed method for a fair comparison with our finetuned LLaMA-2-7B. Prompt template is provided in Table \ref{table:template_chatgpt_baseline}.
    \item \textbf{ChatIE \cite{wei2023zero}}: ChatIE is a framework that transforms the zero-shot event detection task into a multi-turn question-answering problem. Here, LLMs are first prompted (as shown in Table \ref{table:template_chatie_baseline} )to identify the event type and then sequentially prompted to identify the trigger.  
    \item \textbf{\citet{gao2023exploring}}: This work explores the feasibility of ChatGPT as a zero-shot event datection model and further analyses the impact of event definitions, in-context examples and counterfactual examples in the prompt template in Table \ref{table:template_guo_baseline}. We prompt ChatGPT with event definitions and positive examples in our implementation as this setup performed best on \citet{gao2023exploring} evaluation. 
\end{itemize}

\begin{table*}
\centering
\begin{tabular}{p{0.9\linewidth}}
\toprule
\textbf{Event Type:}\\
ceasefire\\
\\

\midrule
\textbf{Event Ontology:}\\
Parent: historical\_event \\
Sons: ceasefire, coronation, rebellion, revolt\\
\\

\midrule
\textbf{Event Definition:} \\
Definition 1: A ceasefire is a temporary cessation of armed conflict marked by an agreement between warring factions to lay down their weapons and cease all hostile activities, often in pursuit of a negotiated settlement or peace agreement.\\
\\
Definition 2: Ceasefire is a legal agreement between two or more conflicting parties to temporarily halt hostilities, usually to allow humanitarian aid to reach the affected civilian population or to negotiate a long-term peace agreement.\\
\\
Definition 3: A ceasefire refers to a state of truce or temporary peace between warring factions, allowing time for diplomatic negotiations and discussions to take place in pursuit of a more sustainable cessation of violence. \\
...\\
Definition 9: Ceasefire denotes a moment of respite in fighting between belligerent groups, often created through negotiations, that allows for the provision of humanitarian aid and the establishment of channels for peacebuilding and reconciliation.\\
\\
Definition 10: Ceasefire is a crucial tool in conflict resolution and peacebuilding, serving as a vital step toward addressing underlying conflicts and arriving at a more permanent peace settlement. \\
\\

\midrule
\textbf{Event Samples:} \\
Sample 1: In 1991, the Persian Gulf War ended with a ceasefire. \\
Trigger: ceasefire \\
\\
Sample 2: The two warring factions in the region agreed on a temporary ceasefire to allow humanitarian aid to reach the affected areas. \\
Trigger: ceasefire \\
\\
Sample 3: After weeks of intense fighting, the UN brokered a ceasefire between the government and rebel forces. \\
Trigger: ceasefire \\
...\\
Sample 9: The military forces of two countries agreed to a ceasefire to allow for the exchange of prisoners of war.\\
Trigger: ceasefire \\
\\
Sample 10: The two neighboring countries agreed to a ceasefire to de-escalate tensions and engage in peace talks. \\
Trigger: ceasefire \\
\\
\bottomrule
\end{tabular}
\caption{Examples for the generated data for event ceasefire.}
\label{table:example_GenData_1}
\end{table*}

\begin{table*}
\centering
\begin{tabular}{p{0.98\linewidth}}
\toprule
\textbf{Event Type:}\\
Change\_event\_time\\
\\

\midrule
\textbf{Event Ontology:}\\
Parent: Change\_event\_time \\
Sons: Holding\_off\_on, Change\_event\_duration\\
\\

\midrule
\textbf{Event Definition:} \\
Definition 1: A Change\_event\_duration is an event where the original duration of an activity or event is modified, either by increasing or decreasing the allotted time, to ensure the completion of the task or event.\\
\\
Definition 2: Change\_event\_duration is an event that entails modifying the estimated duration of a particular activity or event based on assessment or evaluation data, such as delays, technical difficulties, or resource constraints.\\
\\
Definition 3: Change\_event\_duration refers to the event of making revisions to the originally planned duration of an activity or event, typically done to accommodate changing priorities, shifting schedules, or other external factors. \\
...\\
Definition 9: A Change\_event\_duration is an event that involves adjusting the length of time allocated for a particular activity or event, motivated by a need to optimize efficiency, manage resources, or meet project objectives.\\
\\
Definition 10: Change\_event\_duration refers to the event of extending or reducing the time frame for executing a particular task or activity, often done to accommodate shifting business needs or changing stakeholder demands. \\
\\

\midrule
\textbf{Event Samples:} \\
Sample 1: The concert promoters extended the length of the show due to popular demand. \\
Trigger: extended \\
\\
Sample 2: The conference organizers shortened the duration of the keynote speeches to accommodate more panel discussions. \\
Trigger: shortened \\
\\
Sample 3: The wedding planner adjusted the ceremony start time to avoid overlapping with the sunset. \\
Trigger: adjusted \\
...\\
Sample 9: The film festival prolonged its run for an extra day to showcase more entries.\\
Trigger: prolonged \\
\\
Sample 10: The charity event shortened its fundraising campaign due to unexpected budget cuts. \\
Trigger: shortened \\
\\
\bottomrule
\end{tabular}
\caption{Examples for the generated data for event Change\_event\_time.}
\label{table:example_GenData_2}
\end{table*}

\begin{table*}
\centering
\small
\begin{tabular}{p{0.98\linewidth}}
\toprule
\textbf{Ontology:} Parent: Arriving; Sons:  Visiting scenario arrival, Drop in on, Access scenario\\
\midrule
\textbf{Parent: } Arriving\\
\textbf{Event Definition 1: } The act of Arriving involves the physical or virtual arrival at a destination or location, often involving anticipation and preparation for the event or activity that will follow.\\
...\\
\textbf{Event Definition 10: } The Arrival event captures the moment when someone arrives at a particular location, often involving an emotional and physical shift as they transition into a new environment.\\
\textbf{Sample 1: } The school field trip participants \textbf{arrived} at the museum and were greeted by the tour guide.\\
...\\
\textbf{Sample 10: } The visitors \textbf{arrived} at the aquarium and were led to the dolphin show by the staff.\\
\midrule
\textbf{Son 1:} Drop in on\\
\textbf{Event Definition 1: } Drop in on refers to an unplanned and impromptu visit to a friend or acquaintance, often characterized by a surprise element and lack of formal invitations or arrangements.\\
...\\
\textbf{Event Definition 10: } The act of drop in on signifies an unscheduled visit to an individual's place without prior notice or appointment, possibly to offer support or check on their well-being.\\
\textbf{Sample 1: } Sarah decided to \textbf{pop in} on her friend who lived nearby and catch up.\\
...\\
\textbf{Sample 10: } Jane had some free time on her hands and wanted to \textbf{pay a visit} to her former college roommate.\\
\midrule
\textbf{Son 2:} Visiting scenario arrival\\
\textbf{Event Definition 1: } Visiting scenario arrival entails arriving at a planned destination, such as a theater or concert, where specific events have been organized for the visitor's entertainment or education, creating a unique and memorable experience.\\
...\\
\textbf{Event Definition 10: } The event of Visiting scenario arrival involves arriving at a location designated for a pre-planned gathering, such as a family reunion, where participants come together to socialize, network, or reconnect.\\
\textbf{Sample 1: } The investors \textbf{arrived} at the company's headquarters for their business presentation.\\
...\\
\textbf{Sample 10: } The family \textbf{reached} the theme park with their pre-booked ride tickets.\\
\bottomrule
\end{tabular}
\caption{Qualitative examples of DivED dataset. DivED contains diverse sibling events, high-quality samples with diverse triggers for each event type. The event definitions significantly distinguish the slight differences between sibling events.}
\label{table:qualitative_examples_full}
\end{table*}

\end{document}